\documentclass[10pt,twocolumn]{article} 
\usepackage{simpleConference}

\usepackage{amsmath,amssymb,amsfonts}
\usepackage{mathrsfs}
\usepackage{algorithm,algpseudocode}
\usepackage{eucal}
\usepackage{bbm}
\usepackage{tabularx,booktabs}
\newcolumntype{P}[1]{>{\centering\arraybackslash}p{#1}}
\newcolumntype{M}[1]{>{\centering\arraybackslash}m{#1}}

\usepackage{bm}
\usepackage{multicol}
\usepackage{multirow}
\usepackage{subfigure}

\usepackage{colortbl}
\usepackage{arydshln}
\usepackage{bbding}
\usepackage{graphicx}
\usepackage{textcomp}
\usepackage{algorithmicx}
\usepackage{algpseudocode}
\usepackage{float}
\usepackage{times}
\usepackage{graphicx}
\usepackage{amssymb}
\usepackage{url}
\usepackage[hidelinks,colorlinks=true,linkcolor=blue,citecolor=blue]{hyperref}

\usepackage[small]{caption}
\usepackage{cite}
\usepackage{amsmath,amssymb,amsfonts}
\usepackage{mathrsfs}
\usepackage{algorithm,algpseudocode}
\usepackage{eucal}
\usepackage{bbm}
\usepackage{bm}
\usepackage{multicol}
\usepackage{multirow}
\usepackage{subfigure}
\usepackage[dvipsnames]{xcolor}
\usepackage{colortbl}
\usepackage{arydshln}
\usepackage{bbding}
\usepackage{graphicx}
\usepackage{textcomp}
\usepackage{float}

\newcommand{\etal}[1]{\emph{et al}.}

\usepackage[capitalize]{cleveref}

\crefname{section}{Sec.}{Secs.}
\Crefname{section}{Section}{Sections}
\Crefname{table}{Table}{Tables}
\crefname{table}{Tab.}{Tabs.}

\algnewcommand{\LineComment}[1]{\State \(\triangleright\) #1}

\begin{document}

\title{MedSapiens: Taking a Pose to Rethink Medical Imaging Landmark Detection}

\author{
Marawan~Elbatel$^{1}$\footnotemark[1],
Anbang~Wang$^{1}$\footnotemark[1],
Keyuan~Liu$^{2}$,
Kaouther~Mouheb$^{3}$,
Enrique~Almar\text{-}Munoz$^{4}$, \\
Lizhuo~Lin$^{2}$,
Yanqi~Yang$^{2}$,
Karim~Lekadir$^{5}$,
Xiaomeng~Li$^{1}$
\\[6pt]
$^{1}$The Hong Kong University of Science and Technology, Hong Kong \\
$^{2}$The University of Hong Kong, Hong Kong \\
$^{3}$Erasmus MC, Netherlands \\
$^{4}$Medical University of Innsbruck, Austria \\
$^{5}$University of Barcelona, Spain
}

\maketitle
\thispagestyle{empty}
\begin{abstract}
This paper does not introduce a novel architecture; instead, it revisits a fundamental yet overlooked baseline: adapting human-centric foundation models for anatomical landmark detection in medical imaging. While landmark detection has traditionally relied on domain-specific models, the emergence of large-scale pre-trained vision models presents new opportunities. In this study, we investigate the adaptation of \textbf{Sapiens}, a human-centric foundation model designed for pose estimation, for medical imaging through multi-dataset pretraining, establishing a new state-of-the-art across multiple datasets. Our proposed model, \textbf{MedSapiens}, demonstrates that human-centric foundation models, inherently optimized for spatial pose localization, provide strong priors for anatomical landmark detection, yet this potential has remained largely untapped. We benchmark \textbf{MedSapiens} against existing state-of-the-art models, achieving up to \textbf{5.26\%} improvement over generalist models and up to \textbf{21.81\%} improvement over specialist models in the average success detection rate (SDR). To further assess \textbf{MedSapiens}' adaptability to novel downstream tasks with few annotations, we evaluate its performance in limited-data settings, achieving~\textbf{2.69\%} improvement over the few-shot state-of-the-art in the SDR metric. Code and model weights are available at~\url{https://github.com/xmed-lab/MedSapiens}.
\end{abstract}

\footnotetext[1]{Equal contribution.}

\begin{figure*}[t]
    \centering
    \includegraphics[width=\textwidth]{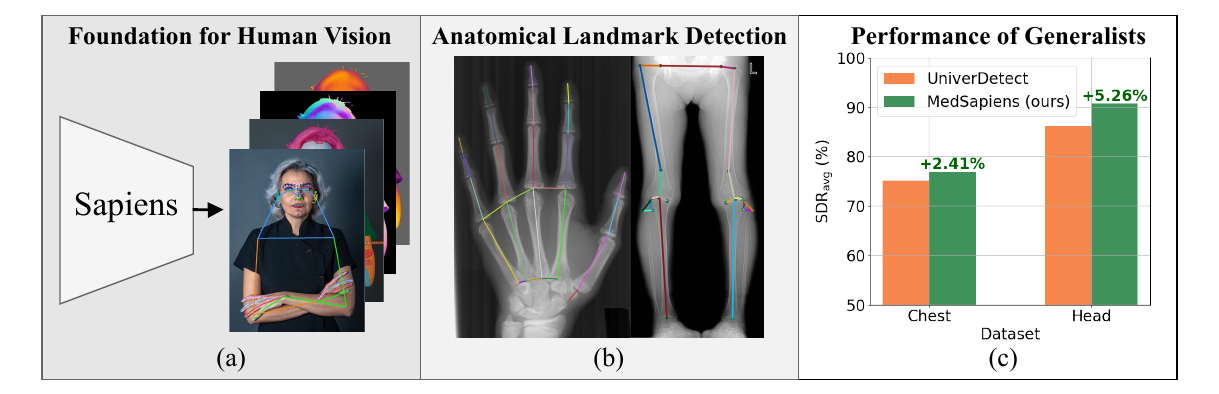}
-    \caption{ (a) Sapiens~\cite{khirodkar2024sapiens}, a foundation model devised for vision-centric tasks such as pose estimation. (b) Anatomical landmark detection shares a hierarchical structure with human-centric tasks. (c) By adapting Sapiens for anatomical landmark detection,  MedSapiens surpasses the existing SOTA generalist model UniverDetect~\cite{univerdetect2024_landmark}.}
    \label{fig:intro_fig}
\end{figure*}

\section{Introduction}

Anatomical landmark detection refers to identifying specific points on anatomical structures, essential for spatial understanding and guiding clinical tasks such as diagnostics, treatment planning, and surgical navigation. Accurate landmark detection is crucial in clinical practice for applications such as brain tumor resection~\cite{brain_mri_ultrasound_landmark_detection_miccai2024}, infant hip dysplasia diagnosis~\cite{ultrasound_diganosis_BHI_2024_landmark,GCN_miccai_2024_hip_topological_landmark}, and cephalometric analysis in orthodontics~\cite{miccai_2024_cephalometric_protoypical_landmark}. Current methods for anatomical landmark detection are predominantly single-task oriented, constrained by limited dataset sizes, and exhibit poor generalization to novel tasks. While foundation models have demonstrated improvements in accuracy and generalization for segmentation~\cite{zhang2023customized,zhao2024biomedparse,med_clip_sam}, classification~\cite{classification_foundation}, and registration~\cite{tian2024unigradicon_foundation_reg,demir2024multigradicon_fountation_reg}, their progress in anatomical landmark detection remains constrained by the lack of a unified framework that jointly optimizes multiple downstream tasks. Therefore, developing an anatomical landmark foundation model is crucial to improving generalization, enabling cross-task adaptability, and achieving broader clinical impact in medical imaging.


Existing approaches for anatomical landmark detection leverage a variety of priors, including geometric priors~\cite{miccai_2024_geomertric_prompt_landmark_detection}, generative priors~\cite{TMI24_generative_prior_landmark}, and anatomical priors~\cite{GCN_miccai_2024_hip_topological_landmark}. These methods, applied in fully supervised scenarios, employ strategies such as prototypical learning~\cite{miccai_2024_cephalometric_protoypical_landmark}, contrastive learning~\cite{brain_mri_ultrasound_landmark_detection_miccai2024}, generative modeling~\cite{sashimi_miccai_2024_ddpm_landmark}, multi-resolution learning~\cite{miccai_2023_landmark_multiresolution_hybrid_network}, as well as regularization techniques~\cite{miccai_2024_graph_landmark_detection}. Additionally, few-shot settings have been explored through multi-domain pre-training~\cite{UOD_miccai_2023_landmark}, or leveraging existing ImageNet-pre-trained foundation models~\cite{Mia_FMOSD_MICCAI2024}. A few works aimed to achieve broader generalization abilities, as demonstrated by GU2Net~\cite{gu2net,universal_model_BME_landmark}, and UniverDetect~\cite{univerdetect2024_landmark}. Nevertheless, these approaches face significant limitations due to pre-training on the constrained size of landmark detection datasets, which typically comprise only a few hundred images. Furthermore, these methods fail to fully exploit the potential of publicly available foundation models, leading to suboptimal generalization performance on novel tasks (See~\Cref{tab:novel_task_few_shot}).

Foundation models offer strong generalization for segmentation and classification and have gained widespread adoption in both generic computer vision and medical imaging~\cite{wu2023self,zhang2023customized,zhao2024biomedparse,med_clip_sam,classification_foundation,jiang2025detectpointprediction, elbatel2024fd}. However, adapting foundation models for anatomical landmark detection is not as straightforward. Unlike segmentation and classification, which have well-established pre-trained backbones, landmark detection lacks direct counterparts, making transfer learning less effective.

Sapiens~\cite{khirodkar2024sapiens}, a recent foundation model pre-trained on over \textit{300 million} in-the-wild images, sets a new scale benchmark for pose estimation. Pose estimation shares conceptual similarities with anatomical landmark detection, as they aim to capture spatial hierarchies and contextual relationships between key points.

To this end, we demonstrate that \textit{efficiently fine-tuning foundation models devised for pose estimation can achieve state-of-the-art (SOTA) performance across diverse medical imaging landmark detection tasks}. Building on this insight, we introduce \textbf{MedSapiens}, a foundation model trained on diverse anatomical landmarks from multiple medical imaging datasets, achieving a new SOTA.

To demonstrate the generalization capability of \textbf{MedSapiens}, we evaluate it on few-shot samples from an unseen novel task, where \textit{MedSapiens outperforms the SOTA, achieving a 2.69\% improvement in the average success detection rate}. While this may not be entirely surprising given that existing foundation models devised for pose estimation are inherently designed to localize anatomical landmarks in human figures, it has been largely overlooked in the literature, with no prior work explicitly exploring this direction. Our contributions can be summarized as follows:  

\begin{itemize}
    \item We highlight the previously unexplored potential of leveraging human-centric foundation models for anatomical landmark detection in medical imaging.
    \item We introduce \textbf{MedSapiens}, a foundation model for anatomical landmark detection, which outperforms the existing SOTA generalist model with improvements of up to \textbf{5.26\%} in the average success detection rate. When further specialized for a specific task, {MedSapiens} surpasses the SOTA specialist model, achieving improvements of up to \textbf{21.81\%} in the average success detection rate.

\item Finally, to assess the generalization of our model, we evaluate \textbf{MedSapiens} on a \textbf{novel unseen task}, achieving improvements of up to \textbf{2.69\%} over the few-shot SOTA, demonstrating superior cross-task adaptability and generalization capabilities. 
\end{itemize}

\section{Methodology}

\begin{figure*}[t]
    \centering
    \includegraphics[width=\textwidth]{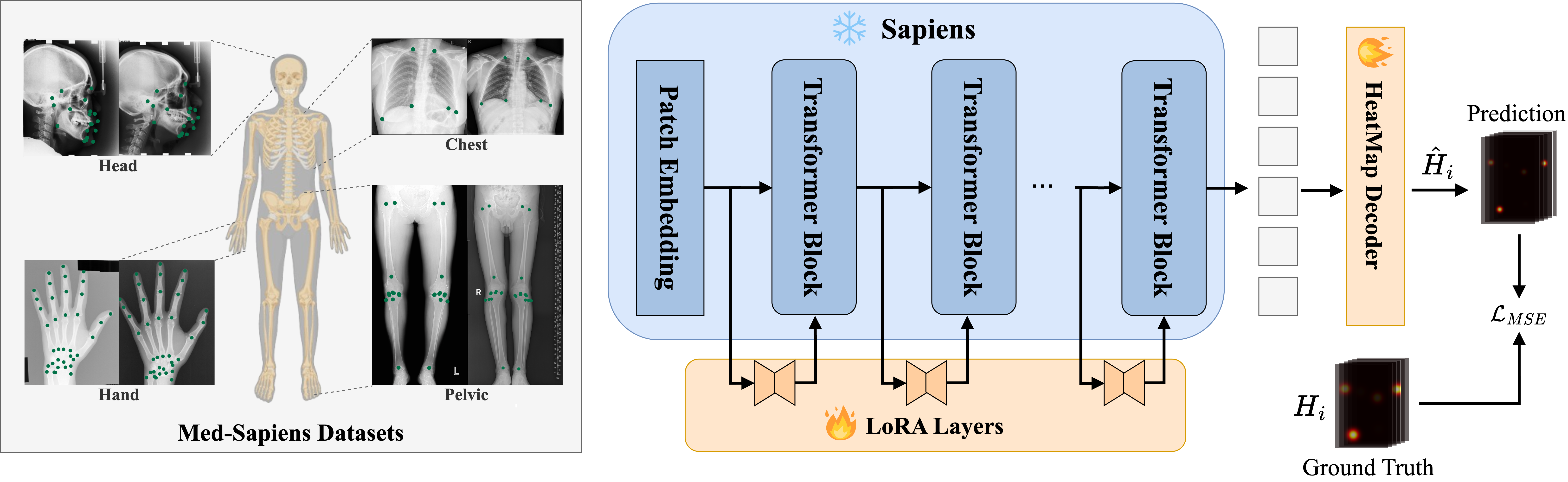}
    \caption{Overall framework for MedSapiens.}
    \label{fig:framework}
    \vspace{-0.5cm}
\end{figure*}

\subsection{Preliminaries}

\noindent{\textbf{Problem Formulation.}}  
Anatomical landmark detection is fundamental in medical imaging, playing a crucial role in diagnosis, surgical planning, and treatment monitoring. The objective is to accurately localize a predefined set of anatomical landmarks within a given medical image. Given an image \( I \in \mathbb{R}^{H \times W} \), the task is formulated as predicting \( N \) landmarks, represented as a set of coordinate pairs \( \{(x_i, y_i)\}_{i=1}^{N} \), where \( (x_i, y_i) \) denote the spatial location of the \( i \)-th landmark. The problem is inherently challenging due to the limited availability of annotated medical datasets, substantial anatomical variability across imaging tasks, and the need to model complex spatial relationships among landmarks.

\noindent{\textbf{MedSapiens Landmark Datasets.}}  
To facilitate robust pretraining of foundation models, we have harmonized a diverse set of publicly available medical landmark datasets spanning different anatomical regions, comprising four datasets with 47,847 annotated landmark points across 1,778 images. The Head X-ray dataset~\cite{head_1,head_2} consists of 400 lateral cephalograms with 19 annotated landmarks and a resolution of 0.1 mm, commonly used in orthodontics. The Hand X-ray dataset includes 909 radiographs with 37 annotated landmarks, normalized to a wrist width of 50 mm introduced in~\cite{hand_introduced} and annotated by~\cite{hand_annotated}. The Chest X-ray dataset~\cite{gu2net} contains 279 images from the China Set, excluding abnormal lungs, with six manually annotated landmarks defining lung boundaries. The BMPLE dataset consists of 190 leg X-rays with 26 annotated landmarks for lower extremity analysis, introduced in CC2D~\cite{yao2022relativedistancemattersoneshot_BLMPE}.


\subsection{Overall Framework}
We propose MedSapiens, a framework designed to leverage large-scale human-centric pretraining for anatomical landmark detection in medical imaging. Built upon the Sapiens model~\cite{khirodkar2024sapiens}, a vision transformer (ViT) trained for human-centric tasks, we adapt it to localize anatomical landmarks across diverse medical imaging modalities. To bridge the gap between human-centric tasks and domain-specific anatomical structures, we pre-train the model on a harmonized collection of public medical landmark datasets, ensuring broad anatomical representation. Given the disparity in dataset scales and the limited availability of annotated landmark detection datasets, full fine-tuning is computationally expensive and prone to overfitting. To allow the model to retain the spatial hierarchies and contextual relationships
from large-scale human-centric pretraining, while effectively adapting to the constraints of medical imaging, we employ Low-Rank Adaptation (LoRA)~\cite{hu2022lora}, a parameter-efficient fine-tuning approach that injects trainable low-rank updates into the transformer's self-attention and projection layers while preserving the pre-trained backbone. Finally, we incorporate a heatmap-based prediction head to refine the extracted features and generate spatial confidence maps, enabling precise and robust landmark localization.

\subsection{Heatmap-Based Decoding}

The Heatmap Head in MedSapiens serves as the decoding mechanism that translates feature representations from the backbone into spatial confidence maps, enabling precise localization of anatomical landmarks. 

Given feature maps \( F \in \mathbb{R}^{h \times w \times C} \) from the transformer backbone, the Heatmap Head applies a series of transposed convolutional layers to progressively upsample the features, followed by \( 1 \times 1 \) convolutions for refinement. The output is a set of heatmaps \( \hat{H} \in \mathbb{R}^{h' \times w' \times N} \), where \( N \) is the number of anatomical landmarks. Each heatmap represents the spatial confidence of a specific landmark.

To supervise the model, \textit{Keypoint Mean Squared Error (MSE)} loss is employed, which compares the predicted heatmaps \( \hat{H} \) to the ground-truth heatmaps \( H \). Each ground-truth heatmap is generated by placing a Gaussian kernel centered at the true landmark location. The MSE loss is then defined as:
\[
\mathcal{L}_{\text{MSE}} = \frac{1}{N} \sum_{i=1}^N \| \hat{H}_i - H_i \|_2^2,
\]
where \( \hat{H}_i \) and \( H_i \) represent the predicted and ground-truth heatmaps for the \( i \)-th landmark, respectively. 

\section{Experiments}

\begin{table*}[t]
\centering
\caption{Comparisons with SOTA methods on the Head and Hand datasets.}
\label{tab:comparison}
\resizebox{2.0\columnwidth}{!}{
\begin{tabular}{l|ccc|cc|ccc|cc}
\hline
 & \multicolumn{5}{c|}{\textbf{Hand Dataset}} & \multicolumn{5}{c}{\textbf{Head Dataset}} \\
\hline
& 
\multicolumn{3}{c|}{SDR (\%)↑} & \multirow{2}{*}{\(\text{SDR}_{\text{avg}}\)} & \multirow{2}{*}{MRE(mm)↓} 

&\multicolumn{3}{c|}{SDR (\%)↑} & \multirow{2}{*}{\(\text{SDR}_{\text{avg}}\)} & \multirow{2}{*}{MRE(mm)↓}

\\
\cline{2-4} \cline{7-9}
&2 mm & 4 mm & 10 mm & & 

& 2 mm & 3 mm & 4 mm & &

\\
\hline

  \multicolumn{9}{c}{Generalist model} 
  \\
\hline

UniverDetect~\cite{univerdetect2024_landmark}  

& 95.76   & 99.30   & {99.91} &98.32& $0.71_{\pm1.78 }$ 

& 75.87   & 88.35   & 94.59   & 86.27&$1.55_{\pm1.74}$

\\

\textbf{MedSapiens (ours)} 

& 95.87 & 99.70 & 100.0 &\textbf{98.52}&{$\mathbf{0.664_{\pm.110}}$} 

& 82.29 & 92.80 & {97.33} & \textbf{90.81}&$\mathbf{1.275_{\pm.285}}$

\\

\hline
  \multicolumn{9}{c}{Specialist model} 
  \\
\hline

NFDP~\cite{TMI24_generative_prior_landmark}

& 95.11 & 99.44 & 99.97 & 98.17&$0.673_{\pm.152}$ 

& 82.00 & 92.44 & 96.99 & 90.48&{$1.245_{\pm.276}$} 

\\

\textbf{MedSapiens w/ LoRA (ours)}

& {96.32} & {99.74} & {100.0} &\textbf{98.68}&
$\mathbf{0.638_{\pm.106}}$ 

& {83.14} & {92.88} & 97.09 & \textbf{91.04}
&$\mathbf{1.244_{\pm.276}}$

\\


\bottomrule

\end{tabular}
}
\end{table*}

\noindent \textbf{Baselines.}
Given the extensive methods proposed over the past years for anatomical landmark detection, we focus on comparing MedSapiens with the two most recent SOTA approaches: {NFDP}\cite{TMI24_generative_prior_landmark} and {UniverDetect}\cite{univerdetect2024_landmark}. NFDP employs generative priors to model anatomical distributions~\cite{TMI24_generative_prior_landmark}, while UniverDetect leverages a category-agnostic framework to ensure cross-domain generalizability~\cite{univerdetect2024_landmark}. Since UniverDetect~\cite{univerdetect2024_landmark} is not publicly available, we follow its training, validation, and testing dataset splits, which are predefined for each dataset, and reproduce NFDP and MedSapiens to ensure a fair comparison. To evaluate generalization, we test MedSapiens in few-shot settings on a novel unseen dataset during training, LDTeeth~\cite{WanAnb_GeometricGuided_MICCAI2025}, comparing it to diverse baselines, including GU2Net~\cite{gu2net}, FM-OSD~\cite{Mia_FMOSD_MICCAI2024}, and GeoSapiens~\cite{WanAnb_GeometricGuided_MICCAI2025}. For GU2Net~\cite{gu2net}, we expand training by incorporating the unseen task dataset alongside its original head, hand, and chest datasets. For FM-OSD~\cite{Mia_FMOSD_MICCAI2024}, we adapt its one-shot framework for few-shot learning by increasing the training data with additional samples and using a multi-sample approach for template matching.

\noindent \textbf{Implementation Details.} The \textbf{generalist} {MedSapiens} leverages the Sapiens 0.3B Vision Transformer as the backbone, pre-trained on over 300 million in the wild--images~\cite{khirodkar2024sapiens}. Fine-tuning is achieved using Low-Rank Adaptation, applied to the \(qkv\) and projection layers of the transformer with a rank of 4. The model is trained using the AdamW optimizer with an initial learning rate of $5 \times 10^{-4}$ and a layer-wise decay rate of 0.85. To ensure robustness, random flips, photometric distortions, and coarse dropout are incorporated as data augmentations. MedSapiens employs a top-down pose estimation pipeline for training and evaluation, assuming the bounding box encompasses the full image. To adapt MedSapiens to different datasets as a \textbf{specialist}, we further fine-tune it on each specific dataset using LoRA~\cite{hu2022lora}, referred to as \textbf{MedSapiens w/ LoRA} in our analysis. LoRA
is selected because it imposes minimal architectural changes,
provides parameter-efficient adaptation, and preserves the
generalist model’s representations while enabling effective
specialization.

\noindent{\textbf{Evaluation Metrics.}} We strictly follow existing baselines~\cite{TMI24_generative_prior_landmark,univerdetect2024_landmark} for the evaluation protocol. Performance is evaluated using Mean Radial Error (MRE) and Successful Detection Rate (SDR) at various thresholds on the original scale. We follow prior work’s normalization strategies for datasets without physical spacing information (e.g., hand dataset wrist width is assumed to be 50 mm)~\cite{TMI24_generative_prior_landmark,univerdetect2024_landmark}.

\subsection{Quantitative Results}

\begin{figure}[h]
    \centering
\includegraphics[width=\linewidth]{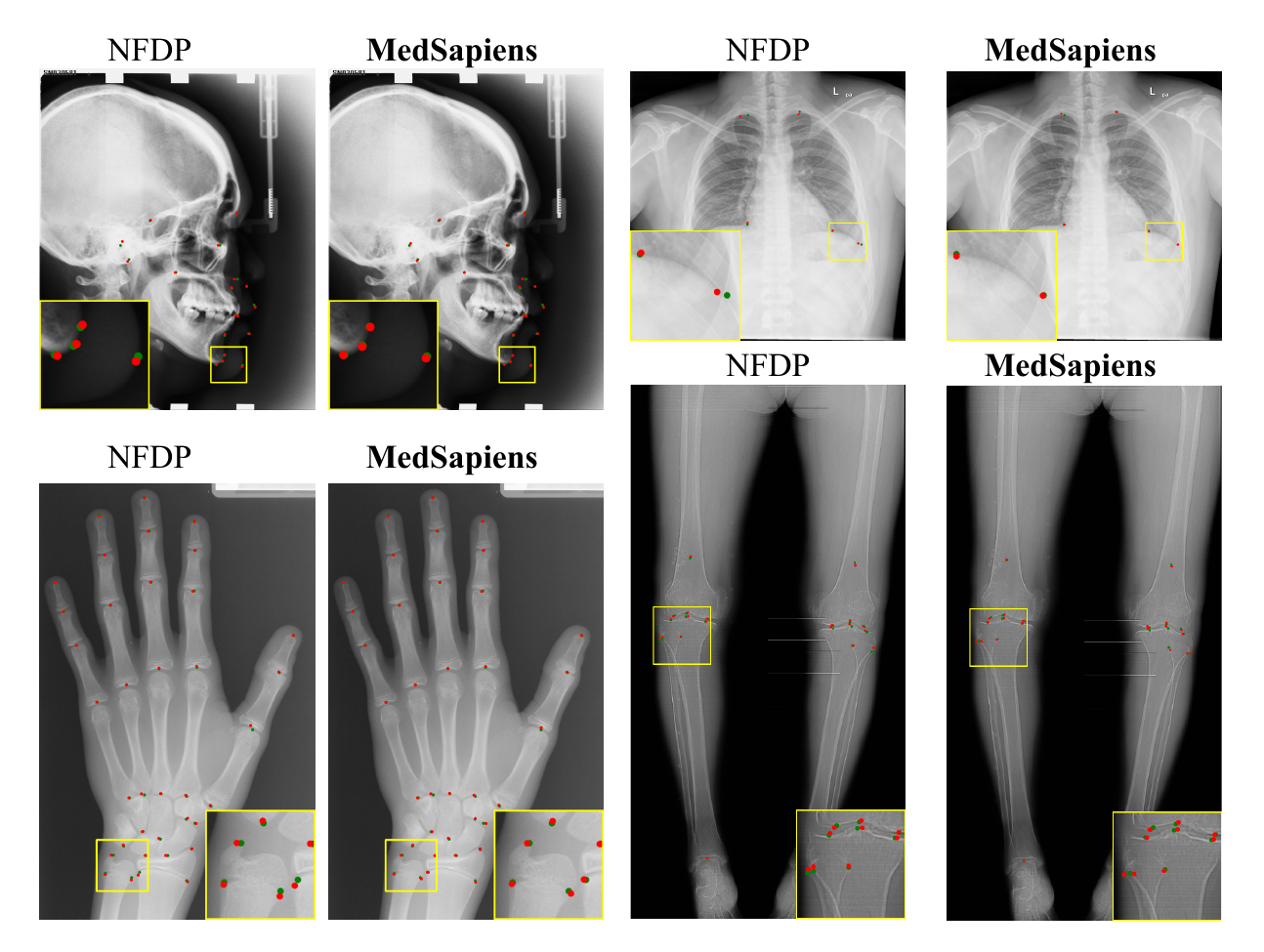}
    \caption{Qualitative results of NFDP and MedSapiens on the four testing datasets, where red points indicate predicted landmarks and
green points represent ground-truth labels.}
\label{fig:medsapiens_nfdp_qualitative}
\vspace{-0.5cm}
\end{figure}

\Cref{tab:comparison} and \Cref{tab:comparison_chest_legs} show comparisons of {MedSapiens} with SOTA methods on the {Hand}, {Head}, {Legs}, and {Chest} datasets. The results demonstrate that {MedSapiens} outperforms existing methods. Compared to the generalist model, UniverDetect~\cite{univerdetect2024_landmark}, MedSapiens achieves relative \textbf{improvements} of 0.20\%, \textbf{5.26\%}, and \textbf{2.41\%} on the average SDR across datasets, respectively for the hand, head, and chest datasets. Specifically, MedSapiens surpasses UniverDetect with relative \textbf{improvements} of 0.11\%, \textbf{8.46\%}, and \textbf{6.28\%} on the strict metric of 2mm, respectively for the hand, head, and chest datasets, offering precise anatomical landmark detection. Given that UniverDetect~\cite{univerdetect2024_landmark} does not release its model weights by the time of submission and is trained and evaluated on privately held datasets, we were unable to compare it to the publicly available Legs dataset.

\begin{table*}[h]
\centering
\caption{Comparisons with SOTA methods on the Chest and Legs datasets. {‡} Results were not reported by the authors, and the model is not publicly available. We follow UniverDetect~\cite{univerdetect2024_landmark} testing splits, which are predefined for each dataset, and
reproduce NFDP~\cite{TMI24_generative_prior_landmark} to ensure a fair comparison. }
\label{tab:comparison_chest_legs}
\resizebox{2.0\columnwidth}{!}{
\begin{tabular}{l|ccc|cc|ccc|cc}
\hline
 & \multicolumn{5}{c|}{\textbf{Legs Dataset}} & \multicolumn{5}{c}{\textbf{Chest Dataset}} \\
\hline
& \multicolumn{3}{c|}{SDR (\%)↑} & \multirow{2}{*}{\(\text{SDR}_{\text{avg}}\)} & \multirow{2}{*}{MRE(mm)↓} & \multicolumn{3}{c|}{SDR (\%)↑} & \multirow{2}{*}{\(\text{SDR}_{\text{avg}}\)} & \multirow{2}{*}{MRE(px)↓} 
\\
\cline{2-4} \cline{7-9}

& 2 mm & 4 mm & 10 mm & & 

& 3 px & 6 px & 9 px & &

\\
\hline

  \multicolumn{11}{c}{Generalist model} 
  \\
\hline
UniverDetect~\cite{univerdetect2024_landmark} 

& {‡} & {‡}& {‡} & {‡} & {‡} 

& 50.81 & {82.52} & 92.27 & 75.19& $4.06_{\pm3.73}$ 

\\

\textbf{MedSapiens (ours)} 

& 48.08 & 85.08 & 97.15 & {76.77}&$2.69_{\pm.555}$ 

& 54.00 & 83.67 & 93.33 &\textbf{77.00}&$\mathbf{3.715_{\pm1.31}}$

\\

\hline
  \multicolumn{11}{c}{Specialist model} 
  \\

\hline

NFDP~\cite{TMI24_generative_prior_landmark}  
& 50.69 & 84.62 & 96.92 &75.89& $2.685_{\pm.617}$ 

& 31.00 & 70.00 & 89.67 & 63.55& $5.13_{\pm1.44}$

\\

\textbf{MedSapiens w/ LoRA (ours)}
& 53.54 & 87.77 & 97.54 & \textbf{79.61}&$\mathbf{2.509_{\pm.556}}$

& 51.67 & 82.33 & 93.67 & \textbf{77.41}&$\mathbf{3.734_{\pm1.24}}$

\\
\bottomrule
\end{tabular}
}
\end{table*}

\begin{table}[t]
\centering
\caption{Comparison with SOTA Few-shot methods on the Downstream Task.}
\label{tab:novel_task_few_shot}
\resizebox{\columnwidth}{!}{
\begin{tabular}{l|ccc|c|c}

\hline
& \multicolumn{3}{c|}{SDR (\%)↑} & \multirow{2}{*}{\(\text{SDR}_{\text{avg}}\)} & \multirow{2}{*}{MRE(px)↓} 
\\
\cline{2-4} 
    & 0.5mm & 1mm & 2mm & &\\
\hline

GU2Net~\cite{gu2net} & 45.21 & 64.12 & 81.90 & 63.74 & 1.312 \\
FM-OSD~\cite{Mia_FMOSD_MICCAI2024} & 32.95 & 55.79 & 77.62 & 55.45 & 1.520 \\
NFDP~\cite{TMI24_generative_prior_landmark} & 55.01 & 80.40 & 92.09 & 75.83 & 0.825 \\

\hline
\multicolumn{6}{c}{Taking a Pose} 
\\
\hline
Sapiens (Baseline) & 60.07 & 82.72 & 93.13 & 78.64 & 0.766 \\
GeoSapiens~\cite{WanAnb_GeometricGuided_MICCAI2025} & {63.19} & {84.14} & {93.36} & {80.23} & {0.747} \\
\textbf{Med-Sapiens (ours)} & {65.66} & {87.31} & {94.21} & \textbf{82.39} & \textbf{0.724} \\
\hline
\end{tabular}}
\end{table}

\begin{figure*}[h]
    \centering
    \includegraphics[width=\linewidth]{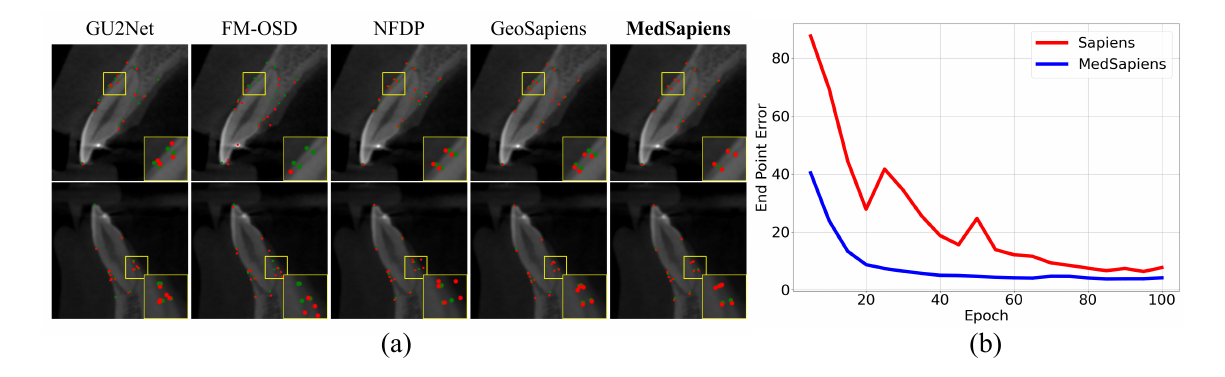}
    \vspace{-0.5cm}
    \caption{Qualitative comparison of our MedSapiens method with existing baselines: (a)  Results on dental images, where red points indicate predicted landmarks and green points represent ground-truth labels. Our MedSapiens achieves superior alignment with the ground truth compared to other methods.
(b) Convergence analysis of MedSapiens vs. Sapiens in terms of end-point error across epochs. Our method exhibits faster and more stable convergence with lower error.
}
\label{fig:few_shot_qualitative_results}
\end{figure*}

While a generalist model offers broad transferability across diverse downstream tasks, task-level specialization remains desirable. To obtain such \textbf{specialist} models, we fine-tune the generalist MedSapiens on each target dataset using LoRA~\cite{hu2022lora}, denoted as {MedSapiens w/ LoRA}. We demonstrate that MedSapiens w/ LoRA can achieve SOTA performance compared to the specialist model, NFDP\cite{TMI24_generative_prior_landmark}, without requiring any prior assumptions or distribution knowledge over the dataset. Specifically, MedSapiens w/ LoRA surpasses NFDP\cite{TMI24_generative_prior_landmark} with relative \textbf{improvements} of 0.52\%, 0.62\%, \textbf{4.90\%}, and \textbf{21.81\%} on the average SDR across datasets, respectively for the hand, head, legs, and chest datasets. Additionally, it achieves relative \textbf{improvements} of 1.27\%, 1.39\%, \textbf{5.62\%}, and \textbf{66.68\%} on the strict metric of 2mm, respectively for the hand, head, legs, and chest datasets, offering precise anatomical landmark detection. The substantial improvements observed in the chest X-ray dataset may be attributed to its high anatomical variability, encompassing differences across male and female subjects. Traditional models may struggle due to limited distributional and generative priors, whereas our model demonstrates robustness in adapting to these variations, ensuring more reliable anatomical landmark detection. We present qualitative results in~\cref{fig:medsapiens_nfdp_qualitative} across the four datasets, comparing our method to NFDP.

\subsection{Few-Shot Novel Task: Teeth Landmarks}

We evaluate MedSapiens on a novel task of detecting dental landmarks in Cone-Beam Computed Tomography (CBCT) images under few-shot conditions. Accurate teeth landmark detection, targeting key features such as the cementoenamel junction, physiological crest, and root apex, is vital for precise clinical measurements, which play a crucial role in diagnosis and risk assessment~\cite{teeth_1}. Dentists typically employ strict criteria for detecting landmarks, using a threshold of 0.5 mm for assessing patients with dental diseases due to the necessity for high precision in clinical evaluations~\cite{teeth_1}. We adopt the {LDTeeth} dataset, introduced in GeoSapiens~\cite{WanAnb_GeometricGuided_MICCAI2025}, and follow the same experimental protocol, including its official patient-wise train/test split. Specifically, we use 3 patients for training and 19 patients for testing.

Experimental results on LDTeeth demonstrate that MedSapiens establishes a new SOTA in few-shot dental landmark detection. 
Compared to NFDP~\cite{TMI24_generative_prior_landmark}, MedSapiens improves $SDR_{avg}$ by \textbf{8.65\%}. 
To ensure a fair capacity-matched comparison, we additionally evaluate GeoSapiens~\cite{WanAnb_GeometricGuided_MICCAI2025}, which—similar to MedSapiens—employs a parameter-efficient tuning strategy with the same number of trainable parameters (\textbf{24M}). 
Without relying on complex task-specific geometric loss functions or additional optimization strategies, MedSapiens further achieves a \textbf{+2.69\%} relative gain in $SDR_{avg}$ over GeoSapiens. Moreover, MedSapiens attains higher precision under the strict 0.5\, mm clinical threshold (\textbf{65.66\% vs.\ 63.19\%}), highlighting its stronger anatomical localization capability in clinically relevant scenarios.

\section{Conclusion}
MedSapiens highlights the potential of leveraging large-scale pre-trained models for anatomical landmark detection. By integrating LoRA fine-tuning with a diverse set of anatomical landmark datasets, it outperforms both generalist and specialist benchmarks, achieving state-of-the-art performance. Future work will explore multimodal adaptation and scaling model parameters as larger datasets become available.

\bibliographystyle{abbrv}
\small\bibliography{refs}

\end{document}